\pdfoutput=1

\documentclass[11pt]{article}

\usepackage[preprint]{acl}

\usepackage{times}
\usepackage{latexsym}

\usepackage[T1]{fontenc}

\usepackage[utf8]{inputenc}

\usepackage{microtype}

\usepackage{inconsolata}

\usepackage{graphicx}
\usepackage{tabularx}
\usepackage{multirow}
\usepackage{tablefootnote}
\usepackage{booktabs}
\usepackage{bbm}
\usepackage{enumitem}
\usepackage{float}
\usepackage[hang]{footmisc}
\usepackage{wrapfig}

\usepackage{listings}
\definecolor{codegreen}{rgb}{0,0.6,0}
\definecolor{codegray}{rgb}{0.5,0.5,0.5}
\definecolor{codered}{rgb}{0.8,0.2,0.2}
\lstdefinestyle{mystyle}{
    commentstyle=\color{codegray},
    numberstyle=\color{codegreen},
    stringstyle=\color{codered},
    basicstyle=\ttfamily\footnotesize,
    breakatwhitespace=false,         
    breaklines=true,                 
    captionpos=b,                    
    keepspaces=true,                 
    showspaces=false,                
    showstringspaces=false,
    showtabs=false,                  
    tabsize=2,
    frame=lines,
}

\title{FanOutQA: A Multi-Hop, Multi-Document \\ Question Answering Benchmark for Large Language Models}

\author{Andrew Zhu, \hspace{0.25cm} Alyssa Hwang,  \hspace{0.25cm} Liam Dugan, \hspace{0.25cm} Chris Callison-Burch \\
        University of Pennsylvania \\
        {\tt \{andrz,ahwang16,ldugan,ccb\}@seas.upenn.edu}}

\begin{document}
\maketitle
\begin{abstract}

One type of question that is commonly found in day-to-day scenarios is ``fan-out'' questions, complex multi-hop, multi-document reasoning questions that require finding information about a large number of entities. However, there exist few resources to evaluate this type of question-answering capability among large language models.
To evaluate complex reasoning in LLMs more fully, we present FanOutQA, a high-quality dataset of fan-out question-answer pairs and human-annotated decompositions with English Wikipedia as the knowledge base.
We formulate three benchmark settings across our dataset and benchmark 7 LLMs, including GPT-4, LLaMA 2, Claude-2.1, and Mixtral-8x7B, finding that contemporary models still have room to improve reasoning over inter-document dependencies in a long context. We provide our dataset and open-source tools to run models to encourage evaluation.\footnote{\url{https://fanoutqa.com} \\ \url{https://github.com/zhudotexe/fanoutqa}}

\end{abstract}

\section{Introduction}
In real-world production deployments, large language models (LLMs) are often asked ``fan-out'' questions: questions that require models to find a list of entities and then consult a large number of documents to aggregate information about those entities to answer a user's question. 
This pattern of question can be found commonly in day-to-day scenarios, such as performing a literature review (fan-out over research papers), planning a trip (fan-out over attractions), or choosing where to eat (fan-out over nearby restaurants).
The fan-out task is particularly challenging because it requires multi-hop reasoning across multiple documents, and the combined length of the documents needed to answer the question typically exceeds the length of a model's context window. 
Existing question-answering benchmarks like HotpotQA \cite{yang-etal-2018-hotpotqa}, LongBench \cite{bai2023longbench}, and ZeroSCROLLS \cite{shaham-etal-2023-zeroscrolls} focus on intra-document dependencies or dependencies between a small number of documents, which does not sufficiently evaluate models' performance on this type of task.

\begin{figure}
    \centering
    \includegraphics[width=\columnwidth]{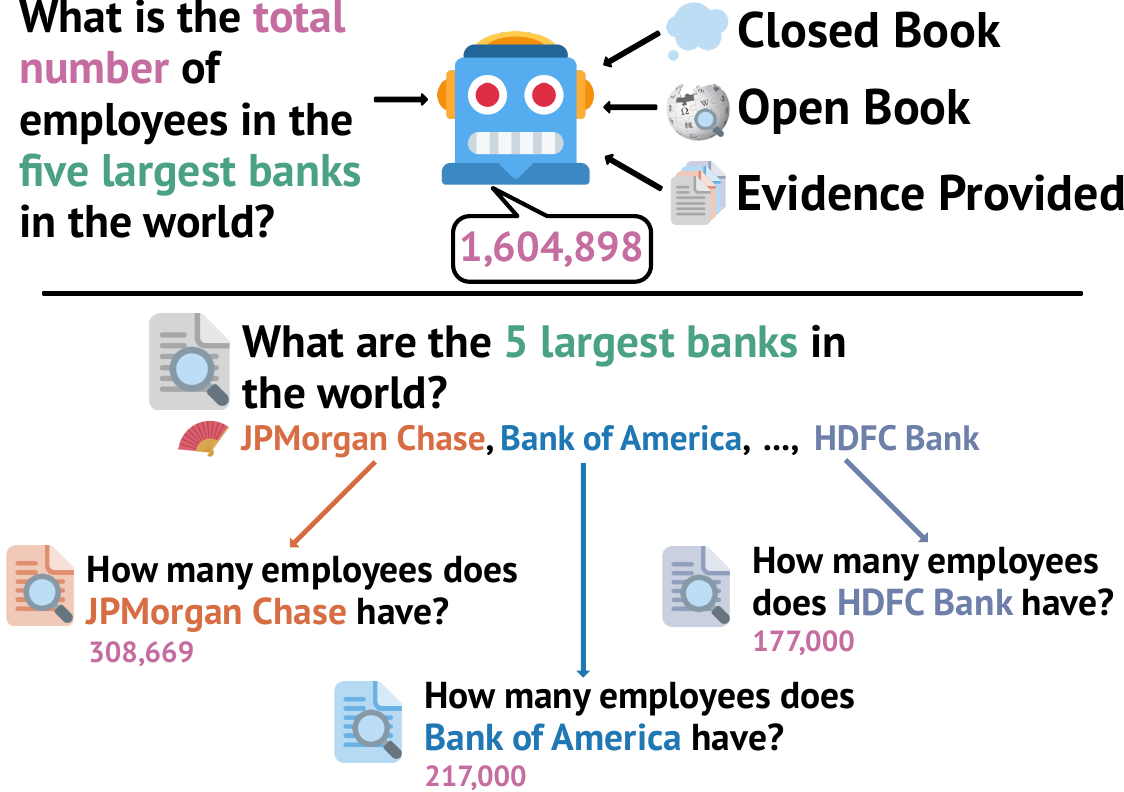}
    \caption{The FanOutQA dataset contains multi-hop, multi-document ``fan-out'' questions along with human-written decompositions (bottom). We formulate three challenge settings for LLMs to answer these fan-out questions to test capabilities of LLMs (top).}
    \label{fig:ex-question}
\end{figure}

In this paper, we present FanOutQA, a high quality dataset of 1,034 information seeking questions, 7,305 human-written decompositions, and their answers, along with a multi-hop, multi-document benchmark using English Wikipedia as its knowledge base. Compared to other question-answering benchmarks, FanOutQA requires reasoning over a greater number of documents, with its main focus being on the fan-out style of question (Figure \ref{fig:ex-question}). 

We formulate three distinct challenge settings over the dataset. The \textbf{closed-book} setting requires the model to answer fan-out questions without external knowledge, testing its general knowledge. The \textbf{open-book} setting gives models access to retrieval tools, testing their ability to retrieve relevant articles and reason across multiple long documents. Finally, the \textbf{evidence-provided} setting provides the models with relevant articles, testing their long-context and multi-hop reasoning capabilities.

We find that the closed- and open-book settings are difficult for modern systems, with the best performing models scoring below 50\%. 
In the open-book setting, retrieved documents outgrow models' context lengths.
In the evidence-provided setting, models' performance correlates strongly with their context length. Human volunteers completing the open-book task score 85\% accuracy, showing room to improve LLM systems.

\section{Related Work}

\paragraph{Multi-Hop Question Answering.} 
HotpotQA \cite{yang-etal-2018-hotpotqa} focuses on using bridge entities to introduce a ``hop'', requiring models to retrieve information about two related entities. 
ComplexWebQuestions \cite{talmor-berant-2018-web} composes simpler questions to create two-hop questions with a similar bridge entity.
2WikiMultiHopQA \cite{ho-etal-2020-constructing} uses manually curated templates to generate two to four-hop questions among entities in the same class.
MuSiQue \cite{trivedi-etal-2022-musique-citefix} presents algorithmically generated questions with nonlinear reasoning chains, which require up to four hops per question.
These datasets focus on simple reasoning chains, with a maximum of four hops.
In FanOutQA, we require nonlinear reasoning chains that are longer than previous multi-hop QA datasets (an average of seven hops per question).

\paragraph{Long Context Evaluations.} 
LongBench \cite{bai2023longbench} is a collection of multiple long-context tasks. In its multi-document QA setting, it builds on top of the multi-hop QA benchmarks discussed above, adding distractor spans to create artificial long documents which are provided to the model. However, it has been shown that this approach does not necessarily increase the complexity of the QA task \cite{min-etal-2019-compositional}.
The Qasper \cite{dasigi-etal-2021-dataset} and SCROLLS \cite{shaham-etal-2022-scrolls} benchmarks present QA tasks that focus primarily on reading comprehension within a single document, rather than reasoning across multiple documents. These benchmarks and others also evaluate different aspects of long context reasoning through subjective summarization tasks \cite{kwan2023m4le} or text span reordering \cite{shaham-etal-2023-zeroscrolls, li2023loogle}, which is beyond the focus of our benchmark.
Unlike previous benchmarks, our open-book setting requires models to \textit{retrieve} and reason over multiple natural long documents (\textit{multi-hop multi-document}), and our evidence-provided setting requires models to perform inter-document reasoning over multiple provided documents. On average, questions in FanOutQA are paired with 172k tokens of evidence spanning 7 documents.

\section{FanOutQA Dataset}
FanOutQA consists of three parts: questions, answers, and evidence. Each question includes a decomposition into sub-questions that can be answered with a single Wikipedia article. The answers to the sub-questions can then be combined to answer the top-level question. We provide these sub-questions, answers, and associated Wikipedia articles as an additional resource for decomposing complex queries. We provide sample questions in Appendix \ref{sec:appendix-examples}, and the dataset's topic distribution in Appendix \ref{sec:appendix-topics}.

\subsection{Dataset Creation}
To create FanOutQA, we recruited 379 undergraduate and graduate students enrolled in AI or NLP courses at a US university to write questions and answers in the fan-out style. 
We required each question to reference at least five different Wikipedia articles to find its answer.
We also tasked the students to decompose their top-level questions into sub-questions, each providing an answer from a single article. The questions were written in a period of one week, ending on November 20, 2023. We stored a snapshot of Wikipedia on the last day to preserve the knowledge source, which we provide with the dataset. We provided a Jupyter notebook to help with writing (see Appendix \ref{sec:appendix-human}) and offered students extra credit for their contributions.

The students produced 1,418 sets of top-level questions, sub-questions, and Wikipedia references. After our filtering pipeline (Appendix \ref{sec:appendix-pipeline}) to ensure the quality of our dataset, we arrive at 1,034 top-level questions and 7,305 sub-questions, across 4,121 distinct Wikipedia articles. We split the dataset into dev and test splits at a ratio of 30\% dev (310), 70\% test (724). We release the full questions, decomposition, and answers of the dev questions, and only the top-level question and list of articles used in the decomposition for the test questions. We maintain a leaderboard of performance on the test set on our website\footnote{\url{https://fanoutqa.com/leaderboard/}}, with a standard submission for generations on the test set.

\subsection{Settings}
We present three different benchmark settings over the data to evaluate different aspects of LLM systems, which we present in order of expected difficulty (most-to-least difficult).

\paragraph{Closed Book.} In what could be considered the most difficult setting, the model is given only the top-level question and must answer it based solely on the knowledge encoded in its parameters. This setting primarily tests the model's general knowledge and establishes a model-specific baseline.

\paragraph{Open Book.} The open book setting gives the model access to the Wikipedia knowledge base along with the top-level question. Using retrieval tools, it can query our dated snapshot of Wikipedia for relevant information across multiple rounds of interaction. Since the questions in FanOutQA require multiple reasoning steps over specific information across a large number of documents, the open book setting is suitable for evaluating retrieval-augmented generation, multi-hop reasoning, and long-horizon question answering.

\paragraph{Evidence Provided.} In this setting, the model is given the top-level question and the text of each Wikipedia article used in the decomposition. 
The model can answer based on information fully within its context window, which evaluates long-context and long-dependency reasoning similar to \citet{li2023loogle}. It can alternatively retrieve the necessary information from the given documents as a simpler retrieval task.

\section{Benchmarking Study}

We benchmarked seven large language models on FanOutQA: GPT-4, GPT-4-turbo, GPT-3.5-turbo, LLaMA 2 70B Chat, Mistral-7B, Mixtral-8x7B, and Claude 2 (more details in Appendix \ref{sec:appendix-models}).
All models generated text with greedy decoding; all local models were run with FP16 precision. 

\begin{figure*}[t]
    \centering
    \includegraphics[width=\textwidth]{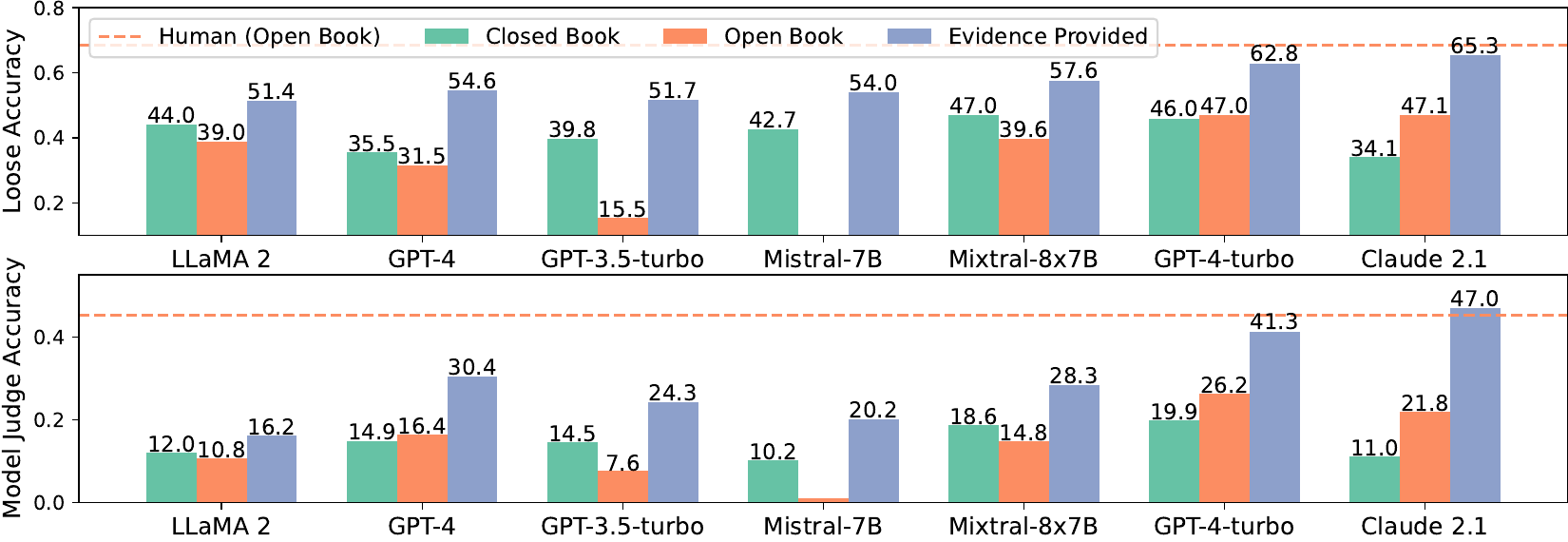}
    \caption{Loose string accuracy and model judged accuracy of all benchmarked models in all settings, including baseline human performance in the open-book setting. See Appendix \ref{sec:appendix-results} for additional metrics.}
    \label{fig:results}
\end{figure*}

\subsection{Metrics}
We report benchmark performance with four classes of metrics.

The first is string accuracy, which we compute after lemmatizing and removing stop words and punctuation from each sequence:
\begin{equation} \label{eq:loose}
    Loose(R, g) = \frac{\sum_{r\in R} \mathbbm{1} [substr(r, g) ] }{|R|}
\end{equation}
\begin{equation} \label{eq:strict}
    Strict(R, g) = \mathbbm{1} [Loose(R, g) = 1]
\end{equation}

Where $R$ is the list of normalized reference answer strings for a given question and $g$ is the normalized candidate generation for that question.

We report the mean proportion of reference answer strings found in the generation (``loose'' accuracy, Eqn. \ref{eq:loose}) and proportion of questions in which \textit{every} answer string was found in the generation (``strict'' accuracy, Eqn. \ref{eq:strict}).

We also report ROUGE-1, ROUGE-2, and ROUGE-L F1-scores \cite{lin-2004-rouge} and BLEURT \cite{sellam-etal-2020-bleurt} scores, consistent with existing related work. 
Finally, we use GPT-4 (\texttt{gpt-4-0613}) to estimate the factual equivalence of the generated and reference answers for each question (prompt in Appendix \ref{sec:appendix-prompts}). We observe that this method is more robust to misspellings and string substitutions, such as ``two'' and ``2'' or ``1 trillion'' and ``1000 billion.'' We present loose string accuracy and the model judge score across all settings in Figure \ref{fig:results}, and tabulate all other results in Appendix \ref{sec:appendix-results}.

\subsection{Closed Book Results}
Using only knowledge encoded in their parameters, models' loose string accuracy ranged from 0.341 (Claude) to 0.470 (Mixtral), with none reaching our estimated human baseline of 0.685 or upper bound of 0.847 (see Section \ref{sec:human_perf_results}). 

Most errors were plausible but incorrect hallucinations. For example, when asked ``which of the top five best selling video games does not feature physical combat,'' GPT-4-turbo answered ``Minecraft'' even though the true answer is Tetris. 




A substantial proportion of errors were unique to OpenAI's GPT models. These models often refused to answer,  citing lack of real time data. Of the models, GPT-4-turbo refused to answer 5\% of the time, GPT-3.5-turbo 10\%, and GPT-4 44\%. 

\subsection{Open Book Results}
\label{sec:open_book_results}
We used Kani \citep{zhu-etal-2023-kani} to provide access to Wikipedia using native function calling (OpenAI's GPT models) or through a structured search query. We split each retrieved document into 1024-character chunks, preferring to split at paragraph and sentence boundaries. We ranked the chunks with a BM25+ \citep{bm25plus} retriever and provided up to half the model's context length of tokens per document. Mistral-7B suffered from severe neural text degeneration \citep{Holtzman2020The} and entered infinite loops when attempting to search, so we omit its open-book results.

Perhaps surprisingly, most models performed worse in the open-book setting than in the closed book setting. We find this to be because models in this setting ``forgot'' the original question as their context windows filled with long retrieved passages across multiple retrieval rounds, outputting a summary of the last retrieved passage instead of answering the question. This is supported by a moderate positive correlation between maximum context window sizes and model-judged accuracy ($r^2 = 0.558$). Models with larger context lengths are able to include a greater amount of information in the context and ``forget'' the original question less often as context windows fill up. We ran two additional experiments where we: a) repeated the original question after each retrieval round and b) limited the context window of all models to the smallest of all models to verify these findings, the results of which are tabulated in Appendix \ref{sec:appendix-additional-experiments}. 


\subsection{Evidence Provided Results}
We use the same retrieval scheme as in the open-book setting, providing models as many chunks as would fit each model's context. Performance correlated strongly with maximum context length in this setting ($r^2 = 0.782$), supporting the proposition that the amount \textit{and quality} of information in a model's context affects its ability to answer fan-out questions. This shows that questions in FanOutQA effectively measure long-context reasoning over very long dependencies. 

\subsection{Human Performance}
\label{sec:human_perf_results}
We conducted a human evaluation to create a human baseline and estimate the upper bound of human performance on FanOutQA. 
We recruited 14 volunteers to each answer 10 FanOutQA questions with access to Wikipedia, similar to the open-book setting. On average, humans took 5-15 minutes to answer each question.
In the open-book setting, the humans score significantly higher than our tested models ($p < 0.05$), achieving a loose accuracy of 68.5\% and model-judged accuracy of 45.2\%. This score may seem low, as the model-judged accuracy does not account for partial credit.
As our only automated metric that accounts for partial credit is not robust to typos and equivalent string substitutions, we also manually evaluate the human answers to establish an upper bound of 84.7\%.


\section{Conclusions}
Fan-out question answering presents several challenges for LLMs, including decomposing complex questions into simpler sub-questions, retrieving documents, extracting relevant information, and multi-hop reasoning over a large number of documents. 
We developed a dataset called FanOutQA for this ambitious task in response to the rapidly improving reasoning abilities and context management strategies in large language models, and we formulate three challenge settings over the dataset. 
We benchmarked the performance of seven state-of-the-art models on our challenge settings, and find that the requirement of fan-out question-answering challenges even the long context capabilities of modern models. 
Accuracy correlated with context length in the open book and evidence-provided but not in the closed book settings, suggesting that more information helps performance. The correlation was stronger in the evidence-provided setting, further suggesting that the quality of information matters as well.

In our experiments, our main goal was to evaluate LLMs’ answers to the top-level questions in the three settings we present. As there may be multiple valid decompositions to achieve a final answer, we don’t evaluate on the similarity between the human-written question decompositions and strategies used by LLMs (most relevant in the Open Book setting). However, we would like to highlight its usefulness for imitation learning (e.g. fine-tuning a function-calling-capable model) as a direction for future work. We also encourage exploration of additional decompositional prompting strategies, such as decomposed prompting \cite{khot2023decomposed} and GenDec \cite{wu2024gendec}.

We encourage researchers to use FanOutQA to evaluate new retrieval-augmented models, long-context models, and other novel LLM systems with our open-source resources.\footnote{\url{https://fanoutqa.com} \\ \url{https://github.com/zhudotexe/fanoutqa}}

\section{Ethics Statement}

Our question writers and human evaluators were compensated with extra credit in a class they were taking or digital items of their choice, with intrinsic value equivalent to or greater than the time effort spent on our task. Participants gave informed consent and were aware of the compensation before accepting the tasks. Data we collected from human annotators is IRB exempt under 45 CFR 46.104, category 2. No personal identifying information was collected from human participants, and any references to individuals found in the dataset reference publicly-available information (i.e. Wikipedia pages).

Wikipedia text is available under the Creative Commons Attribution-ShareAlike 4.0 International License (CC BY-SA) license. We release our dataset under the Creative Commons Attribution-ShareAlike 4.0 International (CC BY-SA) license, and our Python package under the MIT license.

\section{Limitations}

Due to the limitations of text-based metrics, most of our metrics are biased towards recall over precision. The ROUGE metrics measure precision, but LLMs can output extraneous text that penalizes precision without affecting the factual content of the question. This led to many models scoring high in recall but low in precision, leading to an on-average lower reported F1 score. Although using GPT-4 as a judge model helps measure the factual equivalence of two answers, this may be prohibitively expensive to scale to many more thousands of samples.

FanOutQA uses content solely from English Wikipedia, making it a monolingual dataset. It may be plausible to create parallel datasets using the same provided Wikipedia pages found in other languages, but we leave creation and verification of this dataset to future work. 

We focus only on information gathering in this dataset since it possesses useful properties:
\begin{enumerate}
    \item The information is factual with a single answer. Domains such as trip planning require qualitative judgment which complicates evaluation.
    \item We are able to leverage Wikipedia’s backlinks API to enforce the fan-out requirement by examining all articles which commonly link to all evidence used by our human annotators.
    \item Researchers using the dataset are easily able to access the source content as it is available on the web, publicly licensed, and widely available globally without specialized setup.
    \item Information gathering from a closed domain (i.e. Wikipedia) allows us to snapshot the entire domain easily regardless of the path taken by human annotators, allowing us to replicate the entire environment faithfully in evaluation trials.
\end{enumerate}

However, ``fan-out'' tasks extend beyond information gathering, and we are interested in using the methods presented here to extend the scope of the dataset to other domains in future work.

\section*{Acknowledgements}

We would like to thank the members of the lab of Chris Callison-Burch for detailed feedback on the contents of this paper and the members of the Northern Lights Province Discord for their participation in our human evaluation. In particular, we would like to thank Bryan Li for his thoughtful suggestions with regards to our human evaluation and other parts of the paper.

This research is supported in part by the Office of the Director of National Intelligence (ODNI), Intelligence Advanced Research Projects Activity (IARPA), via the HIATUS Program contract \#2022-22072200005. The views and conclusions contained herein are those of the authors and should not be interpreted as necessarily representing the official policies, either expressed or implied, of ODNI, IARPA, or the U.S. Government. The U.S. Government is authorized to reproduce and distribute reprints for governmental purposes notwithstanding any copyright annotation therein.

\bibliography{anthology,custom}

\appendix
\onecolumn

\section{Example Questions}
\label{sec:appendix-examples}

In this section, we provide a sample of various questions found in the FanOutQA dataset, along with their human-written decompositions and answers.

\begin{enumerate}
    \raggedright
    \item 
    \textbf{Q:} What is the duration in minutes and seconds of the top 5 songs on the Billboard Year-End Hot 100 singles list of 2022? \\
    \textbf{Decomposition:}
    \begin{enumerate}
        \item \textbf{Q:} What are the top 5 songs on the list of Billboard Year-End Hot 100 singles of 2022? \\
        \textbf{Evidence:} \url{https://en.wikipedia.org/wiki/Billboard_Year-End_Hot_100_singles_of_2022} \\
        \textbf{A:} Heat Waves, As It Was, Stay, Easy on Me, Shivers 

        \item \textbf{Q:} What is the length of Heat Waves? \\
        \textbf{Evidence:} \url{https://en.wikipedia.org/wiki/Heat_Waves} \\
        \textbf{A:} 3:58 

        \item \textbf{Q:} What is the length of As It Was? \\
        \textbf{Evidence:} \url{https://en.wikipedia.org/wiki/As_It_Was} \\
        \textbf{A:} 2:43

        \item \textbf{Q:} What is the length of Stay? \\
        \textbf{Evidence:} \url{https://en.wikipedia.org/wiki/Stay_(The_Kid_Laroi_and_Justin_Bieber_song)} \\
        \textbf{A:} 2:21

        \item \textbf{Q:} What is the length of Easy on Me? \\
        \textbf{Evidence:} \url{https://en.wikipedia.org/wiki/Easy_on_Me} \\
        \textbf{A:} 3:44

        \item \textbf{Q:} What is the length of Shivers? \\
        \textbf{Evidence:} \url{https://en.wikipedia.org/wiki/Shivers_(Ed_Sheeran_song)} \\
        \textbf{A:} 3:27
    \end{enumerate}
    \textbf{A:} \texttt{\{"Heat Waves": "3:58", "As It Was": "2:43", "Stay": "2:21", "Easy on Me": "3:44", "Shivers": "3:27"\}}

    \item 
    \textbf{Q:} What are the ages of the top 5 most followed people on Instagram?\footnote{As of the dataset epoch of Nov 20, 2023. Retrieved documents return the revision as of this date, so answers are consistent over time.} \\
    \textbf{Decomposition:}
    \begin{enumerate}
        \item \textbf{Q:} Who are the top 5 most followed on Instagram? \\
        \textbf{Evidence:} \url{https://en.wikipedia.org/wiki/List_of_most-followed_Instagram_accounts} \\
        \textbf{A:} Cristiano Ronaldo, Lionel Messi, Selena Gomez, Kylie Jenner, Dwayne Johnson

        \item \textbf{Q:} What is the age of Cristiano Ronaldo? \\
        \textbf{Evidence:} \url{https://en.wikipedia.org/wiki/Cristiano_Ronaldo} \\
        \textbf{A:} 38

        \item \textbf{Q:} What is the age of Lionel Messi? \\
        \textbf{Evidence:} \url{https://en.wikipedia.org/wiki/Lionel_Messi} \\
        \textbf{A:} 36

        \item \textbf{Q:} What is the age of Selena Gomez? \\
        \textbf{Evidence:} \url{https://en.wikipedia.org/wiki/Selena_Gomez} \\
        \textbf{A:} 31

        \item \textbf{Q:} What is the age of Kylie Jenner? \\
        \textbf{Evidence:} \url{https://en.wikipedia.org/wiki/Kylie_Jenner} \\
        \textbf{A:} 26

        \item \textbf{Q:} What is the age of Dwayne Johnson? \\
        \textbf{Evidence:} \url{https://en.wikipedia.org/wiki/Dwayne_Johnson} \\
        \textbf{A:} 51
    \end{enumerate}
    \textbf{A:} \texttt{\{
      "Cristiano Ronaldo": 38,
      "Lionel Messi": 36,
      "Selena Gomez": 31,
      "Kylie Jenner": 26,
      "Dwayne Johnson": 51
    \}}

    \item 
    \textbf{Q:} What are the top 4 best-selling mangas of all time and who is the protagonist for each? \\
    \textbf{Decomposition:}
    \begin{enumerate}
        \item \textbf{Q:} What are the top 4 best-selling mangas of all time? \\
        \textbf{Evidence:} \url{https://en.wikipedia.org/wiki/List_of_best-selling_manga} \\
        \textbf{A:} One Piece, Golgo 13, Case Closed / Detective Conan, Dragon Ball

        \item \textbf{Q:} Who is the protagonist of `One Piece'? \\
        \textbf{Evidence:} \url{https://en.wikipedia.org/wiki/One_Piece} \\
        \textbf{A:} Monkey D. Luffy

        \item \textbf{Q:} Who is the protagonist of `Golgo 13'? \\
        \textbf{Evidence:} \url{https://en.wikipedia.org/wiki/Golgo_13} \\
        \textbf{A:} Duke Togo

        \item \textbf{Q:} Who is the protagonist of `Case Closed / Detective Conan'? \\
        \textbf{Evidence:} \url{https://en.wikipedia.org/wiki/Case_Closed} \\
        \textbf{A:} Shinichi Kudo

        \item \textbf{Q:} Who is the protagonist of `Dragon Ball'? \\
        \textbf{Evidence:} \url{https://en.wikipedia.org/wiki/Dragon_Ball_(manga)} \\
        \textbf{A:} Goku
    \end{enumerate}
    \textbf{A:} \texttt{\{
      "One Piece": "Monkey D. Luffy",
      "Golgo 13": "Duke Togo",
      "Case Closed / Detective Conan": "Shinichi Kudo",
      "Dragon Ball": "Goku"
    \}}

    \item 
    \textbf{Q:} Among the Ivy League universities, which four have the lowest endowments and how many Nobel laureates do each of them have? \\
    \textbf{Decomposition:}
    \begin{enumerate}
        \item \textbf{Q:} Which 4 Ivy League universities have the lowest endowment? \\
        \textbf{Evidence:} \url{https://en.wikipedia.org/wiki/Ivy_League} \\
        \textbf{A:} Brown University, Dartmouth College, Cornell University, Columbia University

        \item \textbf{Q:} How many Nobel laureates does Brown University have? \\
        \textbf{Evidence:} \url{https://en.wikipedia.org/wiki/Brown_University} \\
        \textbf{A:} 11

        \item \textbf{Q:} How many Nobel laureates does Dartmouth College have? \\
        \textbf{Evidence:} \url{https://en.wikipedia.org/wiki/Dartmouth_College} \\
        \textbf{A:} 3

        \item \textbf{Q:} How many Nobel laureates does Cornell University have? \\
        \textbf{Evidence:} \url{https://en.wikipedia.org/wiki/Cornell_University} \\
        \textbf{A:} 62

        \item \textbf{Q:} How many Nobel laureates does Columbia University have? \\
        \textbf{Evidence:} \url{https://en.wikipedia.org/wiki/Columbia_University} \\
        \textbf{A:} 103
    \end{enumerate}
    \textbf{A:} \texttt{\{
      "Brown University": 11,
      "Dartmouth College": 3,
      "Cornell University": 62,
      "Columbia University": 103
    \}}

    \item 
    \textbf{Q:} What is the area in square kilometers of the city that hosts the alma mater of all partners of the main actors from `How I Met Your Mother' who eventually hosted the Academy Awards? \\
    \textbf{Decomposition:}
    \begin{enumerate}
        \item \textbf{Q:} Who are the main actors in `How I Met Your Mother'? \\
        \textbf{Evidence:} \url{https://en.wikipedia.org/wiki/How_I_Met_Your_Mother} \\
        \textbf{A:} Josh Radnor, Jason Segel, Cobie Smulders, Neil Patrick Harris, Alyson Hannigan, Cristin Milioti

        \item \textbf{Q:} Which of these actors hosted the Academy Awards? \\
        \textbf{Evidence:} \url{https://en.wikipedia.org/wiki/List_of_Academy_Awards_ceremonies} \\
        \textbf{A:} Neil Patrick Harris

        \item \textbf{Q:} Who is the partner of Neil Patrick Harris? \\
        \textbf{Evidence:} \url{https://en.wikipedia.org/wiki/Neil_Patrick_Harris} \\
        \textbf{A:} David Burtka

        \item \textbf{Q:} What is the alma mater of David Burtka? \\
        \textbf{Evidence:} \url{https://en.wikipedia.org/wiki/David_Burtka} \\
        \textbf{A:} University of Michigan

        \item \textbf{Q:} What city is the University of Michigan in? \\
        \textbf{Evidence:} \url{https://en.wikipedia.org/wiki/University_of_Michigan} \\
        \textbf{A:} Ann Arbor, Michigan

        \item \textbf{Q:} What is the area of the city of Ann Arbor? \\
        \textbf{Evidence:} \url{https://en.wikipedia.org/wiki/Ann_Arbor,_Michigan} \\
        \textbf{A:} 73.35 sq km
    \end{enumerate}
    \textbf{A:} \texttt{73.35 sq km}

    \item 
    \textbf{Q:} What are the five most popular grape varieties from the Bordeaux appellation, and which area of Bordeaux are they most planted in? \\
    \textbf{Decomposition:}
    \begin{enumerate}
        \item \textbf{Q:} What are the five most popular grape varieties from the Bordeaux appellation? \\
        \textbf{Evidence:} \url{https://en.wikipedia.org/wiki/Bordeaux_wine} \\
        \textbf{A:} Cabernet Sauvignon, Cabernet Franc, Merlot, Semillon, Sauvignon Blanc

        \item \textbf{Q:} Which area of Bordeaux is Cabernet Sauvignon most planted in? \\
        \textbf{Evidence:} \url{https://en.wikipedia.org/wiki/Cabernet_Sauvignon} \\
        \textbf{A:} Haut-Medoc

        \item \textbf{Q:} Which area of Bordeaux is Cabernet Franc most planted in? \\
        \textbf{Evidence:} \url{https://en.wikipedia.org/wiki/Cabernet_Franc} \\
        \textbf{A:} Saint-Emilion

        \item \textbf{Q:} Which area of Bordeaux is Merlot most planted in? \\
        \textbf{Evidence:} \url{https://en.wikipedia.org/wiki/Merlot} \\
        \textbf{A:} Saint-Emilion and Pomerol

        \item \textbf{Q:} Which area of Bordeaux is Semillon most planted in? \\
        \textbf{Evidence:} \url{https://en.wikipedia.org/wiki/S%C3%A9millon} \\
        \textbf{A:} Saint-Emilion

        \item \textbf{Q:} Which area of Bordeaux is Sauvignon Blanc most planted in? \\
        \textbf{Evidence:} \url{https://en.wikipedia.org/wiki/Sauvignon_blanc} \\
        \textbf{A:} Pessac-Leognan and Graves
    \end{enumerate}
    \textbf{A:} \texttt{\{
      "Cabernet Sauvignon": "Haut-Medoc",
      "Cabernet Franc": "Saint-Emilion",
      "Merlot": "Saint-Emilion and Pomerol",
      "Semillon": "Saint-Emilion",
      "Sauvignon Blanc": "Pessac-Leognan and Graves"
    \}}
\end{enumerate}

\section{Dataset Topic Distribution}
\label{sec:appendix-topics}

We queried topics by using GPT-4 to suggest a list of associated topics for each question, then manually reviewed the topics and merged similar ones (e.g. ``Film'' and ``Film Studies''). A question may have multiple associated topics. The top 25 topics covered by the questions are tabulated in Table \ref{tab:topics}.

\begin{wraptable}{r}{0.45\textwidth}
    \begin{tabular}{ l|c|c }
    \toprule 
    \textbf{Topic} &    \textbf{\#} & \textbf{\%}  \\
    \midrule															
    Geography               & 345 & 18.22\% \\
History                 & 230 & 12.14\% \\
Sports                  & 166 & 8.76\%  \\
Film Studies            & 101 & 5.33\%  \\
Education               & 94  & 4.96\%  \\
Economics               & 92  & 4.86\%  \\
Politics                & 90  & 4.75\%  \\
Demographics            & 79  & 4.17\%  \\
Business                & 77  & 4.07\%  \\
Music                   & 67  & 3.54\%  \\
Culture                 & 46  & 2.43\%  \\
Statistics              & 41  & 2.16\%  \\
Literature              & 27  & 1.43\%  \\
Video Games             & 25  & 1.32\%  \\
Technology              & 23  & 1.21\%  \\
Television              & 22  & 1.16\%  \\
Linguistics             & 21  & 1.11\%  \\
Architecture            & 20  & 1.06\%  \\
Finance                 & 20  & 1.06\%  \\
Astronomy               & 19  & 1.00\%  \\
International Relations & 15  & 0.79\%  \\
Physics                 & 15  & 0.79\%  \\
Law                     & 14  & 0.74\%  \\
Japanese Culture        & 14  & 0.74\%  \\
Other                   & 231 & 12.20\% \\
    \bottomrule															
    \end{tabular}
    \caption{Breakdown of question topics included in FanOutQA. Each question may be associated with multiple topics.}
    \label{tab:topics}
\end{wraptable}

There is a slight bias towards questions including a Geography or History component likely due to the example questions given to the question writers. We used vector similarity to deduplicate questions, and in our manual review of similar questions ensured that questions explore distinct topics by removing questions that were simple word-edits of each other (in addition to simple duplicates). Although there is a slight bias towards these domains, no one topic dominates the entire dataset, and we do not believe that the bias has a significant impact on the final conclusions.

\section{Filtering Pipeline}
\label{sec:appendix-pipeline}

To assess the quality of our dataset and remove unsuitable questions, we used computational methods to identify candidates for removal and manually reviewed them after each round. We started with a heuristic-based algorithm to flag two common indicators of low-quality questions: top-level answers not being composed of sub-question answers and multiple sub-questions using the same Wikipedia article as evidence. Next, we ensured that the knowledge base was being used appropriately by verifying that each sub-question answer is contained in the referenced article. Since Wikipedia is a large resource and the writers may not have seen every article related to their questions, we used the OpenAI embeddings (\verb|text-embedding-3-small|, henceforth ``embeddings''; \citealp{neelakantan2022text}) of top-level questions and article titles to retrieve the 30 most similar Wikipedia articles for each question. If any of these articles contained all answers to the sub-questions, we removed the entire example from the dataset. This ensures that the questions both can and need to be answered by the fan-out method.

In the final round of reviewing the quality of our dataset, we used GPT-4 (\verb|gpt-4-0613|) with greedy sampling to help remove or fix poorly phrased questions (prompts in Appendix \ref{sec:appendix-prompts}). We prompted GPT-4 to identify if a question is not objective, such as ``What are five inventions in the Industrial Revolution?'' or ``Who are the five most famous celebrities?'' It was also instructed to identify questions that were missing numeric units and suggest grammar corrections. We manually reviewed all LLM-assisted modifications before deduplication. Finally, we considered duplicate questions to have embeddings with cosine similarity within 0.9. We manually reviewed these duplicates and selected one to remain in the final dataset.

\section{Models Used}
\label{sec:appendix-models}

We benchmarked the following state-of-the-art LLMs' performance on FanOutQA. Where needed, the specific model's key/sub-version is provided.

\textbf{Commercial Models}
\begin{itemize}
    \raggedright
    \setlength\itemsep{0em}
    \item GPT-4 (\texttt{gpt-4-0613}, \citealp{openai2023gpt4})
    \item GPT-4-turbo (\texttt{gpt-4-0125-preview}\footnote{\url{https://platform.openai.com/docs/models/gpt-4-and-gpt-4-turbo}})
    \item GPT-3.5-turbo (\texttt{gpt-3.5-turbo-1106}\footnote{\url{https://platform.openai.com/docs/models/gpt-3-5-turbo}})
    \item Claude (\texttt{claude-2.1}\footnote{\url{https://www.anthropic.com/news/claude-2-1}})
\end{itemize}

\textbf{Open-Source Models}
\begin{itemize}
    \raggedright
    \setlength\itemsep{0em}
    \item LLaMA 2 70B Chat (\texttt{Llama-2-70b-chat}, \citealp{touvron2023llama})
    \item Mistral 7B (\texttt{Mistral-7B-Instruct-v0.2}, \citealp{jiang2023mistral})
    \item Mixtral 8x7B (\texttt{Mixtral-8x7B-Instruct-v0.1}, \citealp{jiang2024mixtral})
\end{itemize}

All models were sampled using greedy decoding, and local models were loaded using FP16 precision on 3 NVIDIA RTX A6000s. We provided the seed \texttt{31415} to OpenAI's GPT models for deterministic generation.

\section{Results Table}
\label{sec:appendix-results}

We tabulate the results of each model and metric in Table \ref{tab:results}.

\begin{table}[H]
    \small
    \begin{tabular}{ l|c|c|c|c|c|c|c|c }
\toprule \multicolumn{9}{c}{\bf Closed Book} \\ \midrule															
\textbf{Model} &	\textbf{Ctx Size} &		\textbf{Loose} &	\textbf{Strict} &			\textbf{ROUGE-1} &			\textbf{ROUGE-2} &			\textbf{ROUGE-L} &	\textbf{BLEURT} &	\textbf{GPT Judge} \\
\midrule															
\textbf{LLaMA 2 70B} & 	4,096 & 		0.440 & 	0.058 & 			0.285 & 			0.149 & 			0.238 & 	0.441 & 	0.120 \\
\textbf{GPT-4} & 	8,096 & 		0.355 & 	0.066 & 			0.313 & 			0.177 & 			0.267 & 	0.419 & 	0.149 \\
\textbf{GPT-3.5-turbo} & 	16,384 & 		0.398 & 	0.058 & 			0.401 & 			0.227 & 			0.342 & 	0.455 & 	0.145 \\
\textbf{Mistral-7B} & 	32,768 & 		0.427 & 	0.055 & 			0.260 & 			0.123 & 			0.212 & 	0.449 & 	0.102 \\
\textbf{Mixtral-8x7B} & 	32,768 & 		\textbf{0.470} & 	0.081 & 			0.302 & 			0.158 & 			0.254 & 	0.466 & 	0.186 \\
\textbf{GPT-4-turbo} & 	128,000 & 		0.460 & 	\textbf{0.101} & 			\textbf{0.482} & 			\textbf{0.290} & 			\textbf{0.409} & 	\textbf{0.493} & 	\textbf{0.199} \\
\textbf{Claude 2.1} & 	200,000 & 		0.341 & 	0.041 & 			0.412 & 			0.208 & 			0.344 & 	0.426 & 	0.110 \\
\midrule \multicolumn{9}{c}{\bf Open Book} \\ \midrule															
\textbf{Model} &	\textbf{Ctx Size} &		\textbf{Loose} &	\textbf{Strict} &			\textbf{ROUGE-1} &			\textbf{ROUGE-2} &			\textbf{ROUGE-L} &	\textbf{BLEURT} &	\textbf{GPT Judge} \\
\midrule															
\textbf{LLaMA 2 70B} & 	4,096 & 		0.390 & 	0.064 & 			0.157 & 			0.075 & 			0.131 & 	0.443 & 	0.108 \\
\textbf{GPT-4} & 	8,096 & 		0.315 & 	0.057 & 			0.208 & 			0.106 & 			0.183 & 	0.427 & 	0.164 \\
\textbf{GPT-3.5-turbo} & 	16,384 & 		0.155 & 	0.032 & 			0.114 & 			0.051 & 			0.099 & 	0.338 & 	0.076 \\
\textbf{Mistral-7B} & 	32,768 & 		--- &	--- &			--- &			--- &			--- &	--- &	--- \\
\textbf{Mixtral-8x7B} & 	32,768 & 		0.396 & 	0.055 & 			0.173 & 			0.078 & 			0.147 & 	0.449 & 	0.148 \\
\textbf{GPT-4-turbo} & 	128,000 & 		0.470 & 	\textbf{0.109} & 			\textbf{0.356} & 			\textbf{0.207} & 			\textbf{0.314} & 	\textbf{0.487} & 	\textbf{0.262} \\
\textbf{Claude 2.1} & 	200,000 & 		\textbf{0.471} & 	0.086 & 			0.295 & 			0.157 & 			0.253 & 	0.485 & 	0.218 \\
\midrule
\textbf{Human} & --- &		0.685 & 	0.289 & 			0.344 & 			0.210 & 			0.307 & 	0.413 & 	0.452 \\
\midrule  \multicolumn{9}{c}{\bf Evidence Provided} \\ \midrule															
\textbf{Model} &	\textbf{Ctx Size} &		\textbf{Loose} &	\textbf{Strict} &			\textbf{ROUGE-1} &			\textbf{ROUGE-2} &			\textbf{ROUGE-L} &	\textbf{BLEURT} &	\textbf{GPT Judge} \\
\midrule															
\textbf{LLaMA 2 70B} & 	4,096 & 		0.514 & 	0.077 & 			0.376 & 			0.206 & 			0.304 & 	0.472 & 	0.162 \\
\textbf{GPT-4} & 	8,096 & 		0.546 & 	0.144 & 			0.500 & 			0.301 & 			0.413 & 	0.530 & 	0.304 \\
\textbf{GPT-3.5-turbo} & 	16,384 & 		0.517 & 	0.102 & 			0.455 & 			0.252 & 			0.358 & 	0.497 & 	0.243 \\
\textbf{Mistral-7B} & 	32,768 & 		0.540 & 	0.088 & 			0.330 & 			0.172 & 			0.264 & 	0.475 & 	0.202 \\
\textbf{Mixtral-8x7B} & 	32,768 & 		0.576 & 	0.135 & 			0.409 & 			0.231 & 			0.343 & 	0.509 & 	0.283 \\
\textbf{GPT-4-turbo} & 	128,000 & 		0.628 & 	0.192 & 			\textbf{0.614} & 			\textbf{0.395} & 			\textbf{0.523} & 	\textbf{0.581} & 	0.413 \\
\textbf{Claude 2.1} & 	200,000 & 		\textbf{0.653} & 	\textbf{0.215} & 			0.423 & 			0.262 & 			0.354 & 	0.508 & 	\textbf{0.470} \\
\bottomrule 														
\end{tabular}
    \caption{Performance of each model on all metrics and all settings. We include human performance in the open-book setting, and omit Mistral-7B's performance in the open-book setting due to catastrophic neural text degeneration.}
    \label{tab:results}
\end{table}

\section{Additional Experiments}
\label{sec:appendix-additional-experiments}

In this section, we list the results of two additional experiments:
\begin{enumerate}
    \item In the open book and evidence provided settings, we limit the context window of all models to the smallest of all models to verify the correlation between context length and performance.
    \item In the open book setting, we repeat the original question after each retrieval round, to ensure that it is always in the context of the model.
\end{enumerate}

\subsection{Limited Context Length}
In this experiment, we fix the context size of each model to be equal to the shortest model's (4096 tokens) to verify correlations between context length and performance, the results of which we tabulate in Table \ref{tab:results-shortctx}.

\begin{table}[H]
    \small
    \begin{tabular}{ l|c|c|c|c|c|c|c|c }
\toprule \multicolumn{9}{c}{\bf Open Book, Context Limited} \\ \midrule															
\textbf{Model} &	\textbf{Ctx Size} &		\textbf{Loose} &	\textbf{Strict} &			\textbf{ROUGE-1} &			\textbf{ROUGE-2} &			\textbf{ROUGE-L} &	\textbf{BLEURT} &	\textbf{GPT Judge} \\
\midrule															
\textbf{LLaMA 2 70B} & 	4,096 & 		0.423 & 	0.066 & 			0.194 & 			0.095 & 			0.163 & 	0.449 & 	0.113 \\
\textbf{GPT-4} & 	4,096 & 		0.236 & 	0.040 & 			0.151 & 			0.071 & 			0.134 & 	0.395 & 	0.102 \\
\textbf{GPT-3.5-turbo} & 	4,096 & 		0.124 & 	0.023 & 			0.099 & 			0.041 & 			0.087 & 	0.326 & 	0.054 \\
\textbf{Mistral-7B} & 	4,096 & 		--- &	--- &			--- &			--- &			--- &	--- &	--- \\
\textbf{Mixtral-8x7B} & 	4,096 & 		\textbf{0.458} & 	\textbf{0.076} & 			\textbf{0.224} & 			0.105 & 			\textbf{0.192} & 	\textbf{0.465} & 	\textbf{0.160} \\
\textbf{GPT-4-turbo} & 	4,096 & 		0.294 & 	0.051 & 			0.194 & 			0.103 & 			0.169 & 	0.427 & 	0.137 \\
\textbf{Claude 2.1} & 	4,096 & 		0.348 & 	0.055 & 			\textbf{0.224} & 			\textbf{0.113} & 			0.187 & 	0.445 & 	0.140 \\
\midrule \multicolumn{9}{c}{\bf Evidence Provided, Context Limited} \\ \midrule															
\textbf{Model} &	\textbf{Ctx Size} &		\textbf{Loose} &	\textbf{Strict} &			\textbf{ROUGE-1} &			\textbf{ROUGE-2} &			\textbf{ROUGE-L} &	\textbf{BLEURT} &	\textbf{GPT Judge} \\
\midrule															
\textbf{LLaMA 2 70B} & 	4,096 & 		0.514 & 	0.077 & 			\textbf{0.376} & 			0.206 & 			0.304 & 	0.472 & 	0.160 \\
\textbf{GPT-4} & 	4,096 & 		0.380 & 	0.083 & 			0.157 & 			0.075 & 			0.131 & 	0.443 & 	0.184 \\
\textbf{GPT-3.5-turbo} & 	4,096 & 		0.425 & 	0.054 & 			0.208 & 			0.106 & 			0.183 & 	0.427 & 	0.162 \\
\textbf{Mistral-7B} & 	4,096 & 		0.466 & 	0.040 & 			0.114 & 			0.051 & 			0.099 & 	0.338 & 	0.134 \\
\textbf{Mixtral-8x7B} & 	4,096 & 		\textbf{0.525} & 	0.102 & 			0.173 & 			0.078 & 			0.147 & 	0.449 & 	0.229 \\
\textbf{GPT-4-turbo} & 	4,096 & 		0.515 & 	\textbf{0.113} & 			0.356 & 			\textbf{0.207} & 			\textbf{0.314} & 	\textbf{0.487} & 	\textbf{0.250} \\
\textbf{Claude 2.1} & 	4,096 & 		0.490 & 	0.084 & 			0.295 & 			0.157 & 			0.253 & 	0.485 & 	0.189 \\
\bottomrule															
\end{tabular}
    \caption{Performance of each model with a fixed context length on all metrics in the open-book and evidence-provided settings. We omit Mistral-7B's performance in the open-book setting due to catastrophic neural text degeneration.}
    \label{tab:results-shortctx}
\end{table}

\subsection{Repeated Question After Retrieval}
In this experiment, we repeat the original question in the prompt after each retrieval round to attempt to mitigate the model ``forgetting'' the original question. The results are tabulated in Table \ref{tab:results-prompt-last}. We found that in this experiment, if the model performed multiple searches, it would ``forget'' some of the retrieved information rather than the original question. For GPT-4, this caused it to re-run a search for previous information (which in turn caused it to ``forget'' other information and re-run another search, ad infinitum). We set a time limit of 5 minutes for each question, and find that GPT-4 times out in 33.1\% of questions. Among the other two tested models, we see no significant improvement in benchmark performance ($p>0.2$) by repeating the original question after each retrieval round. This suggests that the problem cannot be solved by changing the location of the question in a prompt alone: if more information is retrieved than can fit in a model’s context window, some information will always be truncated.

\begin{table}[H]
    \small
    \begin{tabular}{ l|c|c|c|c|c|c|c|c }
\toprule \multicolumn{9}{c}{\bf Open Book, Question Repeated} \\ \midrule															
\textbf{Model} &	\textbf{Ctx Size} &		\textbf{Loose} &	\textbf{Strict} &			\textbf{ROUGE-1} &			\textbf{ROUGE-2} &			\textbf{ROUGE-L} &	\textbf{BLEURT} &	\textbf{GPT Judge} \\
\midrule	
\textbf{LLaMA 2 70B} & 	4,096 & 		0.431 & 	0.065 & 			0.196 & 			0.097 & 			0.166 & 	0.451 & 	0.110 \\
\textbf{GPT-4} & 	8,096 & 		0.230 & 	0.051 & 			0.190 & 			0.095 & 			0.170 & 	0.339 & 	0.140 \\
\textbf{Mixtral-8x7B} & 	32,768 & 		\textbf{0.465} & 	\textbf{0.081} & 			\textbf{0.223} & 			\textbf{0.105} & 			\textbf{0.191} & 	\textbf{0.466} & 	\textbf{0.170} \\
\bottomrule															
\end{tabular}
    \caption{Performance of three models after repeating the original question after each retrieval on all metrics in the open-book setting.}
    \label{tab:results-prompt-last}
\end{table}

\section{Human Instructions}
\label{sec:appendix-human}

\subsection{Question Writing Instructions}

\textit{We presented the following instructions to students in a Google Colaboratory notebook. To write the questions and their decompositions, students wrote them as a Python dictionary, which the notebook validated the structure of before their submission. The remainder of this section contains the verbatim instructions included in the notebook.}

We are creating a challenge problem for natural language processing systems, where systems have to answer questions that require them to read multiple sources.

Specifically, we're looking at "fan-out" questions - where the question itself is not too long, but to answer it requires first looking up (or being supplied) some list of items, then finding out more details about each item.

Your job is to help us write:

\begin{itemize}
    \item these fan-out questions
    \item strategies to answer the questions you write, with relevant Wikipedia articles linked
    \item reference answers to these questions.
\end{itemize}

You'll be using this Colab notebook to make sure the questions and answers are in the right format. Let's take a look at a couple examples, first:

For example, a very simple fan-out question might be:

\begin{quote}
    What was the population of New York and Los Angeles in 1950?
\end{quote}

In this example, the best strategy to answer this question is to split it once into two questions, "What was the population of New York in 1950?" and "What was the population of Los Angeles in 1950"?

\begin{lstlisting}[language=Python]
# EXAMPLE FORMAT - DO NOT MODIFY
example_q1 = {
  "question": "What was the population of New York and Los Angeles in 1950?",
  "strategy": [
    # each question in here is the same structure recursively!
    # we don't need to here, but subquestions can be broken up even further
    {
      "question": "What was the population of New York in 1950?",
      "evidence": "https://en.wikipedia.org/wiki/Demographic_history_of_New_York_City",
      "answer": 7891957
    },
    {
      "question": "What was the population of Los Angeles in 1950?",
      "evidence": "https://en.wikipedia.org/wiki/Los_Angeles",
      "answer": 1970358
    },
  ],
  "answer": {
    "New York": 7891957,
    "Los Angeles": 1970358
  }
}

validate_question(example_q1, is_demonstration=True)
# END EXAMPLE 1
\end{lstlisting}

We can make this question more complex by making the system look up the list of items rather than providing it in the question:

\begin{quote}
    What was the population in 1950 of the 5 current most populous cities in the United States?
\end{quote}

Now, to answer the question, one has to first look up a list of populous cities in the US (the \textit{strategy}), then fan-out based on that information.

\begin{lstlisting}[language=Python]
# EXAMPLE FORMAT - DO NOT MODIFY
example_q2 = {
  "question": "What was the population in 1950 of the 5 current most populous cities in the United States?",
  # use "strategy" for questions that don't depend on the answers to previous questions
  "strategy": [
    {
      "question": "What are the 5 most populous cities in the United States?",
      "evidence": "https://en.wikipedia.org/wiki/List_of_United_States_cities_by_population",
      "answer": ["New York", "Los Angeles", "Chicago", "Houston", "Phoenix"]
    },
  ],
  # use "then" if sub-questions depend on answers to the questions in "strategy"
  "then": [
    {
      "question": "What was the population of New York in 1950?",
      "evidence": "https://en.wikipedia.org/wiki/Demographic_history_of_New_York_City",
      "answer": 7891957
    },
    {
      "question": "What was the population of Los Angeles in 1950?",
      "evidence": "https://en.wikipedia.org/wiki/Los_Angeles",
      "answer": 1970358
    },
    {
      "question": "What was the population of Chicago in 1950?",
      "evidence": "https://en.wikipedia.org/wiki/Chicago",
      "answer": 3620962
    },
    {
      "question": "What was the population of Houston in 1950?",
      "evidence": "https://en.wikipedia.org/wiki/Houston",
      "answer": 596163
    },
    {
      "question": "What was the population of Phoenix in 1950?",
      "evidence": "https://en.wikipedia.org/wiki/Phoenix,_Arizona",
      "answer": 106818
    },
  ],
  "answer": {
    "New York": 7891957,
    "Los Angeles": 1970358,
    "Chicago": 3620962,
    "Houston": 596163,
    "Phoenix": 106818
  }
}

validate_question(example_q2)
# END EXAMPLE 2
\end{lstlisting}

Let's look at one more example that's a bit more complex. We'll ask the question:

\begin{quote}
    Find the female cabinet members of the current US President. Who are those cabinet members and what city/town were they born in?
\end{quote}

Now, we need to look up quite a bit more information:

\begin{lstlisting}[language=Python]
# EXAMPLE FORMAT - DO NOT MODIFY
example_q3 = {
  "question": "Find the female cabinet members of the current US President. Who are those cabinet members and what city/town were they born in?",
  "strategy": [
    {
      "question": "Who is the current US President?",
      "evidence": "https://en.wikipedia.org/wiki/List_of_presidents_of_the_United_States",
      "answer": "Joe Biden",
    }
  ],
  "then": [
    {
      "question": "Who are the female members of Joe Biden's cabinet and what city/town were they born in?",
      "strategy": [
        {
          "question": "Who are the female members of Joe Biden's cabinet?",
          "evidence": "https://en.wikipedia.org/wiki/Cabinet_of_Joe_Biden",
          "answer": ["Kamala Harris", "Janet Yellen", "Deb Haaland", "Gina Raimondo", "Julie Su", "Marcia Fudge", "Jennifer Granholm"]
        }
      ],
      "then": [
        {
          "question": "What city/town was Kamala Harris born in?",
          "evidence": "https://en.wikipedia.org/wiki/Kamala_Harris",
          "answer": "Oakland, California"
        },
        {
          "question": "What city/town was Janet Yellen born in?",
          "evidence": "https://en.wikipedia.org/wiki/Janet_Yellen",
          "answer": "New York City, New York"
        },
        {
          "question": "What city/town was Deb Haaland born in?",
          "evidence": "https://en.wikipedia.org/wiki/Deb_Haaland",
          "answer": "Winslow, Arizona"
        },
        {
          "question": "What city/town was Gina Raimondo born in?",
          "evidence": "https://en.wikipedia.org/wiki/Gina_Raimondo",
          "answer": "Smithfield, Rhode Island"
        },
        {
          "question": "What city/town was Julie Su born in?",
          "evidence": "https://en.wikipedia.org/wiki/Julie_Su",
          "answer": "Madison, Wisconsin"
        },
        {
          "question": "What city/town was Marcia Fudge born in?",
          "evidence": "https://en.wikipedia.org/wiki/Marcia_Fudge",
          "answer": "Cleveland, Ohio"
        },
        {
          "question": "What city/town was Jennifer Granholm born in?",
          "evidence": "https://en.wikipedia.org/wiki/Jennifer_Granholm",
          "answer": "Vancouver, British Colombia"
        },
      ],
      "answer": {
        "Kamala Harris": "Oakland, California",
        "Janet Yellen": "New York City, New York",
        "Deb Haaland": "Winslow, Arizona",
        "Gina Raimondo": "Smithfield, Rhode Island",
        "Julie Su": "Madison, Wisconsin",
        "Marcia Fudge": "Cleveland, Ohio",
        "Jennifer Granholm": "Vancouver, British Colombia"
      }
    }
  ],
  "answer": {
    "Kamala Harris": "Oakland, California",
    "Janet Yellen": "New York City, New York",
    "Deb Haaland": "Winslow, Arizona",
    "Gina Raimondo": "Smithfield, Rhode Island",
    "Julie Su": "Madison, Wisconsin",
    "Marcia Fudge": "Cleveland, Ohio",
    "Jennifer Granholm": "Vancouver, British Colombia"
  },
}

validate_question(example_q3)
# END EXAMPLE 3
\end{lstlisting}

Now it's up to you to write 1-5 of these questions in the format provided!

The questions can be about any topic where information is available on English Wikipedia - it does not necessarily have to be related to the class. Your evidence should be a link to a single page on English Wikipedia. Try to make your questions fairly diverse and unambiguous (e.g. include the units the answer is expected in, if applicable).

The answer to a top-level question must not be available on a singular Wikipedia article. Your question must require looking at at least 5 Wikipedia articles.

If your question does not validate, please read the error to see what changes are needed.

Use this template for each question/subquestion:

\begin{lstlisting}
{
  "question": "YOUR QUESTION HERE",
  "strategy": [
    # subquestions
  ],
  "then": [
    # more subquestions that depend on answering the questions in "strategy" first (if any)
  ],
  "evidence": "link to wikipedia",  # each subquestion needs evidence to answer it, or a recursive strategy - you should either have evidence or strategy, but not both
  "answer": 0  # can be a dict, list, or primitive value
}
\end{lstlisting}

\textbf{Glossary}

\verb|question| (str): The question to be answered. At the root node, this should not be answerable without breaking it up into smaller subquestions.

\verb|strategy| (list of Question): Subquestions to break the question up into. These shouldn't require looking anything up to ask (e.g. see example 1 vs 2).

\verb|then| (list of Question, optional): Subquestions to ask with the information gathered after answering all the subquestions in strategy, if any are needed.

\verb|evidence| (link to Wikipedia): If question can be answered by information found on a single Wikipedia page, the link to that page.

\verb|answer| (dict, list, or primitive): The final answer to the question, after all subquestions have been answered.

Tip: Either evidence or strategy may be present in a subquestion, but not both. If the answer to a question can be found on a single Wikipedia page, use evidence. If you need to break it up into smaller questions, use strategy (and possibly then).

There might be multiple valid strategies to answer a top-level question; use the one that is most intuitive to you. After writing your question, validate it with \verb|validate_question| and see if it makes sense to read.

\textit{Blank code cells follow for question writing.}

\subsection{Question Answering Instructions}

\textit{We presented the following instructions to volunteers participating in our human evaluation after they gave their informed consent. These instructions imitate the Open Book setting for models.}

Thanks for participating in the FanOutQA human evaluation! You will be given 10 questions, and your task is to answer the questions to the best of your ability.

You may use English Wikipedia (\url{https://en.wikipedia.org/wiki/Main_Page}) to search for Wikipedia articles to help you answer each question. \textbf{Do not use Google or other search engines.} Please record which Wikipedia articles you looked at (whether or not you used the information in the article) to answer the questions.

To answer the questions, please make a copy of this Google doc, and fill in your answers in the spaces below. Once you are finished, please send the document as a PDF to <first author's email>.

\begin{itemize}
    \item Answers do not need to be complete sentences.
    \item Answers do not need to be in a particular format - they will be judged by a human.
    \item Some questions may only require a single answer, others may need a list.
    \item You do not need to finish all 10 questions in a single sitting.
    \item You will be awarded based on the number of questions completed, regardless of whether or not the answer is correct. Please do your best to answer correctly though! You will not be given an award if the answers are obviously low-effort.
\end{itemize}

\textit{A list of ten questions, randomly sampled from the FanOutQA test set per participant, follows.}

\section{LLM Prompts}
\label{sec:appendix-prompts}

\subsection{Subjective Flag}

\begin{lstlisting}
SYSTEM: You are assessing how well a given question can be answered. For each submission, assess whether the provided question can be answered deterministically and objectively at a fixed point in time as of January 2024 given access to appropriate information sources.

USER: [Question]: {question}
***
Can the question be answered in a way that is both deterministic (i.e., the answer has a single unambiguously correct answer) and objective (i.e., the answer is based on factual information and not influenced by personal feelings or opinions) at a given point in time? If the question allows for multiple correct answers, it should not be considered deterministic.
For each question, provide a step-by-step reasoning for your assessment before your conclusion, then print only the single character "Y" or "N" (without quotes or punctuation) on its own line corresponding to the correct answer. At the end, repeat just the letter again by itself on a new line.
\end{lstlisting}

If the model's response ended with the letter "N", we flagged the question for manual review.

\subsection{Grammaticality and Unit Suggestions}

\begin{lstlisting}
SYSTEM: You are assessing how well a given question can be answered. For each question and answer, assess whether the question is grammatical and includes the expected units (if applicable).
If the question does not require any changes, output "No change."
Otherwise, rewrite the question to make it grammatical and include any necessary units without changing the provided answer. Output only the rewrite.
If this is not possible, output the word "FLAG" on its own line, followed by your reasoning.

USER: [Question]: {question}
***
[Answer]: {answer}
\end{lstlisting}

If the model's response began with "FLAG", we recorded the response for manual review. Otherwise, if the model's response was not "No change.", we recorded the suggested rewrite. Afterwards, we manually reviewed all suggestions made by the model.

\subsection{Model Judge}

\begin{lstlisting}
SYSTEM: You are comparing a submitted answer to an expert answer on a given question

USER: [BEGIN DATA]
************
[Question]: {question}
************
[Expert]: {reference}
************
[Submission]: {answer}
************
[END DATA]

Compare the factual content of the submitted answer with the expert answer. Ignore any differences in style, grammar, or punctuation.
The submitted answer may either be a subset or superset of the expert answer, or it may conflict with it. Determine which case applies. First, write out in a step by step manner your reasoning about the factual content to be sure that your conclusion is correct. Avoid simply stating the correct answers at the outset. Then print only the single character "A", "B", "C", "D", "E", or "F" (without quotes or punctuation) on its own line corresponding to the correct answer. At the end, repeat just the letter again by itself on a new line.
(A) The submitted answer is a subset of the expert answer and is fully consistent with it.
(B) The submitted answer is a superset of the expert answer and is fully consistent with it.
(C) The submitted answer contains all the same details as the expert answer.
(D) There is a disagreement between the submitted answer and the expert answer.
(E) The answers differ, but these differences don't matter from the perspective of factuality.
(F) The submitted answer does not answer the question or is otherwise invalid.
\end{lstlisting}

If the model's response ended with the letter "B", "C", or "E", we awarded the answer a score of 1.0. Otherwise, we awarded the answer a score of 0.0.

\subsection{Benchmarks}

\textbf{Closed Book}
\begin{lstlisting}
Answer the following question, and output only your answer. If the answer is a list, output one on each line. Current date: 11-20-2023.

[Question]: {question}
\end{lstlisting}

\textbf{Open Book}

As some models did not have native function calling capabilities, we used a different prompt to instruct these models to output a particular machine-parsable format. For models with native function calling, we used the following function and prompt:

\begin{lstlisting}
def search(query: str):
    """Search Wikipedia for an article with the given title, and get its content. If no such article is found, return similar article names."""
\end{lstlisting}

\begin{lstlisting}
Answer the following question, and output only a function call or your answer. If the answer is a list, output one on each line. Current date: 11-20-2023.

[Question]: {question}
\end{lstlisting}

For models without native function calling, we used the following prompt:

\begin{lstlisting}
You have the ability to search Wikipedia for information. To do so, output a message in the format <search>{YOUR_SEARCH_QUERY}</search> (e.g. `<search>List of states and territories of the United States</search>`).
Answer the following question, and output only your answer or a search, but not both. If the answer is a list, output one on each line. Current date: 11-20-2023.

[Question]: {question}
\end{lstlisting}

\textbf{Evidence Provided}
\begin{lstlisting}
*** BEGIN DATA ***

{evidence_documents}
*** END DATA ***

Answer the following question based on the documents above, and output only your answer. If the answer is a list, output one on each line. Current date: 11-20-2023.

[Question]: {question}
\end{lstlisting}

\end{document}